\documentclass[sigconf, screen]{acmart}

\graphicspath{{figures/}}

\usepackage{amsmath,amssymb,amsfonts}
\usepackage{textcomp}
\usepackage{multirow}
\usepackage{tabularx}
\usepackage{makecell}

\AtBeginDocument{%
  }

\copyrightyear{2026}
\acmYear{2026}
\setcopyright{cc}
\setcctype{by}
\acmConference[MM '26]{Proceedings of the 34th ACM International Conference on Multimedia}{November 10--14, 2026}{Rio de Janeiro, Brazil}
\acmBooktitle{Proceedings of the 34th ACM International Conference on Multimedia (MM '26), November 10--14, 2026, Rio de Janeiro, Brazil}
\acmDOI{10.1145/3767308.3838667}
\acmISBN{979-8-4007-2213-4/2026/11}

\begin{document}

\title{SpeechSense: A Paralinguistic-Focused Dataset for Fine-Grained Speech Sentiment Analysis}


\author{Shicheng Ma}
\affiliation{
  \department{Department of Computer Science and Engineering}
  \institution{The Chinese University of Hong Kong}
  \city{Hong Kong}
  \country{Hong Kong}}
\email{shichengma@cuhk.edu.hk}

\author{Wenqian Cui}
\affiliation{
  \department{Department of Computer Science and Engineering}
  \institution{The Chinese University of Hong Kong}
  \city{Hong Kong}
  \country{Hong Kong}}
\email{wenqian.cui@link.cuhk.edu.hk}

\author{Irwin King}
\affiliation{
  \department{Department of Computer Science and Engineering}
  \institution{The Chinese University of Hong Kong}
  \city{Hong Kong}
  \country{Hong Kong}}
\email{king@cse.cuhk.edu.hk}

\begin{abstract}
Recent advances in AI have revolutionized speech processing, yet effective speech understanding requires discerning not just \textit{what} is said, but \textit{how} it is said. Speech Sentiment Analysis plays a critical role in decoding these paralinguistic cues for diverse real-world applications such as recruitment and customer service. However, existing Speech Sentiment Analysis research faces two primary limitations. First, dominant approaches rely on text-centric pipelines that cascade Automatic Speech Recognition with text analysis. This process inevitably discards essential acoustic features like prosody and tone, failing to capture attitudinal meanings in acoustically ambiguous utterances. Second, current benchmarks suffer from a mismatch in label granularity, prioritizing basic emotions (e.g., happy, sad) over the nuanced interpersonal stances (e.g., confident, impatient) necessary for social sensitivity. To address these limitations, we propose a novel dataset, SpeechSense, for fine-grained speech sentiment analysis. Specifically, we define a specialized 8-class taxonomy of interpersonal stances detectable primarily through prosodic cues beyond lexical content alone. We then construct a curated dataset based on this taxonomy, built from high-fidelity speech synthesis and rigorous human validation. Comprehensive experiments across multi-modal LLMs, text-only LLMs, and speech encoders demonstrate that models with acoustic access consistently outperform text-only baselines. These results empirically validate the primacy of acoustic cues in detecting subtle speaker attitudes, highlighting the necessity of SpeechSense.\footnote{Dataset and supplementary materials: \url{https://github.com/Sher13cked/SpeechSense}}
\end{abstract}




\maketitle

\section{Introduction}
\label{sec:intro}
Recent advances in Artificial Intelligence (AI) have dramatically improved machines' ability to process and understand human speech, enabling applications ranging from automated transcription services \cite{radford2023robust} to end-to-end voice assistants \cite{Cui2025,cui2025turnguide,mtr-duplexbench,wang2025ntpp}. For effective speech understanding, AI models require not only recognizing \textit{what} is said (linguistic information) but also discerning \textit{how} it is said (paralinguistic information) \cite{Cui2025,textpro-slm,audio_reasoning_survey}, as paralinguistic information carries additional meanings beyond the spoken content. A notable task that involves paralinguistic understanding is Speech Sentiment Analysis (SSA), where acoustic cues are analyzed to decipher fine-grained speaker attitudes and affective states. This capability is especially critical in professional scenarios such as online recruitment, customer service, and healthcare \cite{Fagherazzi2021Health, Naim2018Interview, Wang2023Call}. For instance, distinguishing ``confident'' from ``nervous'' delivery during a job interview provides insights that speech content alone misses \cite{degrootWhyVisualVocal1999}.

However, existing SSA research faces two key limitations. \textbf{First, dominant approaches rely heavily on text-centric pipelines.} In these systems, speech is first converted to text via Automatic Speech Recognition (ASR) before applying Text-based Sentiment Analysis (TSA) models \cite{shahReviewNaturalLanguage2023}. This cascaded approach inevitably discards paralinguistic cues, including vocal prosody, tone, pauses, stress, and tempo \cite{schullerParalinguisticsSpeechLanguage2013}. Yet, these acoustic features are often the primary carriers of attitudinal meaning \cite{frickCommunicatingEmotionRole}. For example, the phrase ``We can't wait another decade'' can express either passionate advocacy or impatient irritation, depending entirely on the acoustic delivery. This distinction is lost in the text transcription. Additionally, potential transcription errors in the ASR stage can significantly degrade downstream sentiment analysis performance as well \cite{liSpeechEmotionRecognition2024}.
\textbf{Second, existing research suffers from a mismatch in label granularity.} Most current datasets focus on broad sentiment, such as \textit{positive} or \textit{negative}, or basic emotions like \textit{happy} and \textit{sad} \cite{mohmaddarSpeechDatabasesSpeech2024,cui2024moodloopgp}. Dominant datasets in the field, including IEMOCAP \cite{bussoIEMOCAPInteractiveEmotional2008}, RAVDESS \cite{livingstoneRyersonAudioVisualDatabase2018}, and CREMA-D \cite{caoCREMADCrowdSourcedEmotional2014}, are all predicated on these categorical emotions. 
While useful for basic classification, these labels fail to capture the nuanced interpersonal stances.

To address these limitations, we propose a novel dataset, SpeechSense, for fine-grained speech sentiment detection that prioritizes prosodic features over semantic content. Specifically, we introduce a specialized 8-class label set designed to reflect diverse speaker personalities and states, including Confident, Nervous, Warm, Apathetic, Passionate, Impatient, Sarcastic, and Neutral---all detectable primarily through prosodic 
cues beyond lexical content alone \cite{frickCommunicatingEmotionRole, pon-barryProsodicManifestationsConfidence2008, rockwellVocalFeaturesConversational2007, somaItsNotWhat2023, konigDetectingApathyOlder2019}. To overcome the scarcity of human-recorded speech for these attitudes, we construct the dataset utilizing representative Text-to-Speech (TTS) synthesis to generate prosody-rich samples \cite{schuhmannEmoNetVoiceFineGrainedExpertVerified2025}. These samples are subsequently validated through rigorous human annotation. Finally, we validate our dataset through experiments across multi-modal Large Language Models (LLMs) \cite{xuQwen25OmniTechnicalReport2025}, text-only LLMs \cite{qwen2025qwen25technicalreport}, and speech encoders \cite{radford2023robust, hsu2021hubert, baevski2020wav2vec}. Our results demonstrate that models with acoustic access consistently outperform text-only baselines across all architectures, confirming the primacy of prosodic cues for fine-grained sentiment detection.

Our main contributions are summarized as follows:
\begin{itemize}
  \item We define a fine-grained, paralinguistics-focused sentiment label set that captures nuanced speaker states beyond standard basic emotions.
  \item We introduce a curated dataset combining high-fidelity synthesized speech with human validation, addressing the data scarcity for attitudinal labels.
  \item We validate the dataset across multi-modal LLMs, text-only LLMs, and speech encoders, demonstrating that acoustic access is necessary for fine-grained sentiment detection.
\end{itemize}

\section{Related Work}
\label{sec:related_work}

\subsection{Speech Emotion Recognition Datasets}
SSA builds directly upon Speech Emotion Recognition (SER), which traditionally focuses on identifying basic emotional states. We first review established SER datasets to identify their limitations in capturing complex social attitudes.
The IEMOCAP database \cite{bussoIEMOCAPInteractiveEmotional2008}, containing approximately 12 hours of audiovisual data, remains the most widely used corpus, labeled with categorical emotions such as anger, happiness, sadness, and neutrality. Similarly, RAVDESS \cite{livingstoneRyersonAudioVisualDatabase2018} and CREMA-D \cite{caoCREMADCrowdSourcedEmotional2014} provide acted speech samples across similar basic emotion categories. EMO-DB \cite{burkhardt05b_interspeech} offers high-quality German emotional speech but is restricted to seven classes.
Despite their contributions, these datasets share a critical limitation: they primarily model basic \textit{emotions} rather than nuanced \textit{attitudes} or \textit{interpersonal stances}. 
This gap highlights the need for a new dataset tailored to fine-grained speaker states like confidence or apathy.

\subsection{Methodological Paradigms}
Approaches to SSA generally fall into two paradigms: cascade and end-to-end. The cascade paradigm, dominant in industry applications, treats speech analysis as a two-stage process: ASR converts audio to text, followed by TSA models for sentiment classification \cite{shahReviewNaturalLanguage2023}. While this leverages the semantic reasoning power of LLMs, it suffers from inherent flaws. First, it is vulnerable to error propagation; Li et al. \cite{liSpeechEmotionRecognition2024} demonstrate that Word Error Rate (WER) in ASR transcripts significantly degrade downstream sentiment detection accuracy. Second, the conversion to text irreversibly strips away paralinguistic information—the ``how'' of the message \cite{schullerParalinguisticsSpeechLanguage2013}. Research indicates that prosodic cues (e.g., pitch contours, rhythm) are often independent of, or even contradictory to, linguistic content (e.g., in sarcasm) \cite{rockwellVocalFeaturesConversational2007}.
In contrast, end-to-end approaches process raw audio waveforms or continuous speech representations directly, preserving paralinguistic fidelity. Speech encoders such as Whisper \cite{radford2023robust}, HuBERT \cite{hsu2021hubert}, and Wav2Vec2 \cite{baevski2020wav2vec} learn rich acoustic representations from large-scale audio data and have shown strong performance in speech emotion recognition.
More recently, multi-modal LLMs (e.g., Qwen2.5-Omni \cite{xuQwen25OmniTechnicalReport2025}, GPT-4o) have demonstrated the feasibility of understanding speech directly by integrating acoustic and linguistic information in a unified architecture. However, few studies have specifically optimized these architectures for the fine-grained sentiment detection proposed in this work.

\subsection{Synthetic Data in Speech Processing}
The scarcity of labeled data for specific paralinguistic states has driven research into synthetic data generation. While early SER research relies strictly on human-recorded speech, recent advances in expressive TTS have made synthetic data a viable alternative for training and augmentation. Ma et al. \cite{maLeveragingSpeechPTM2024} empirically validate that SER models trained on high-quality synthesized emotional speech can achieve competitive performance, effectively addressing data scarcity in low-resource emotional classes. Furthermore, very recent works like \textit{EmoNet-Voice} (2025) \cite{schuhmannEmoNetVoiceFineGrainedExpertVerified2025} have successfully utilized expert-verified synthetic speech to create fine-grained emotion benchmarks. Following this trajectory, we leverage a ``Role-Play'' prompting strategy grounded in the Stanislavski method of emotion induction \cite{scherer2003vocal} to achieve the high-fidelity prosodic realization required for complex speaker attitudes.

\section{SpeechSense Dataset}
\label{sec:method}

SpeechSense captures nuanced interpersonal stances rather than the high-intensity basic emotions targeted by existing corpora.

\subsection{Fine-Grained Sentiment Label Set}
\label{sec:label_set}

Unlike basic emotions, our objective is to detect \textit{speaker attitudes} and \textit{interpersonal stances} that are critical for understanding complex social dynamics. Discrete emotion frameworks such as Ekman's basic emotions \cite{ekman1992argument} and Plutchik's wheel \cite{plutchik1980general} were designed to characterize internal affective states rather than communicative stances directed toward an interlocutor. Existing characterizations of interpersonal stance are largely dimensional, and no discrete category set has been standardized for annotating interpersonal stance in speech.
We introduce a fine-grained, 8-class label set derived from established acoustic-psychological research. To highlight their distinctive prosodic signatures, these labels are organized into four comparative groups based on their primary acoustic and interactional attributes: Internal Certainty, High-Energy Valence, Social Connection, and Prosodic Deviation. Rather than adopting an existing scheme directly, we adapt the dimensional principles underlying affective models such as PAD \cite{mehrabian1974approach}---valence, arousal, and dominance---to the interpersonal setting, organizing each group as a contrastive pair that isolates one such distinction. The fourth group additionally captures an orthogonal aspect specific to speech: the alignment between prosody and lexical content. Table~\ref{tab:label_definitions} details the definitions and acoustic signatures for each attitude.

\begin{table*}[tb]
\centering
\small
\caption{The SpeechSense Label Set. The labels are categorized into four attribute groups based on shared acoustic and interactional characteristics to capture fine-grained speaker stances.}
\label{tab:label_definitions}
\renewcommand{\arraystretch}{1.15} 
\begin{tabularx}{\textwidth}{l l X}
\toprule
\textbf{Attribute Group} & \textbf{Label} & \textbf{Acoustic \& Psychological Definition} \\
\midrule

\multirow{2}{*}{\textbf{\makecell[l]{Internal Certainty\\(State of Mind)}}} 
& \textbf{Confident} & Signals dominance and assurance. Acoustically characterized by shorter duration, higher intensity, and decisive falling pitch contours \cite{pon-barryProsodicManifestationsConfidence2008}. \\
& \textbf{Nervous} & Signals anxiety and instability. Manifests through pitch jitter (micro-tremors), higher average pitch, and irregular rhythm \cite{laukkaNervousVoiceAcoustic2008}. \\
\midrule

\multirow{2}{*}{\textbf{\makecell[l]{High-Energy Valence\\(Positive vs. Negative)}}} 
& \textbf{Passionate} & Reflects intense positive engagement (Enthusiasm). Distinguished by an expanded pitch range and dynamic contours \cite{frickCommunicatingEmotionRole}. \\
& \textbf{Impatient} & Reflects irritation or time-pressure (Urgency). Distinguished from Passionate by clipped diction, abrupt stops, and a rigid, staccato rhythm \cite{frickCommunicatingEmotionRole}. \\
\midrule

\multirow{2}{*}{\textbf{\makecell[l]{Social Connection\\(Engagement Level)}}} 
& \textbf{Warm} & Facilitates rapport and empathy. Defined by specific vocal qualities such as a softer timbre, breathy voice quality, and steady pitch stability \cite{somaItsNotWhat2023}. \\
& \textbf{Apathetic} & Reflects detachment and lack of interest. Marked by reduced prosodic variability (monotone), prolonged pauses, and a slower speech rate \cite{konigDetectingApathyOlder2019}. \\
\midrule

\multirow{2}{*}{\textbf{\makecell[l]{Prosodic Deviation\\(Complexity)}}} 
& \textbf{Sarcastic} & A complex attitude where prosody contradicts text. Characterized by a lower mean F0, slower tempo, and a nasal quality relative to the baseline \cite{rockwellVocalFeaturesConversational2007}. \\
& \textbf{Neutral} & Serves as the acoustic baseline or control state. Exhibits standard prosody without significant excursions in pitch, energy, or voice quality \cite{livingstoneRyersonAudioVisualDatabase2018}. \\

\bottomrule
\end{tabularx}
\end{table*}

\subsection{Dataset Construction Pipeline}
\label{sec:pipeline}

\begin{figure}[tb]
\centering
\includegraphics[width=\linewidth]{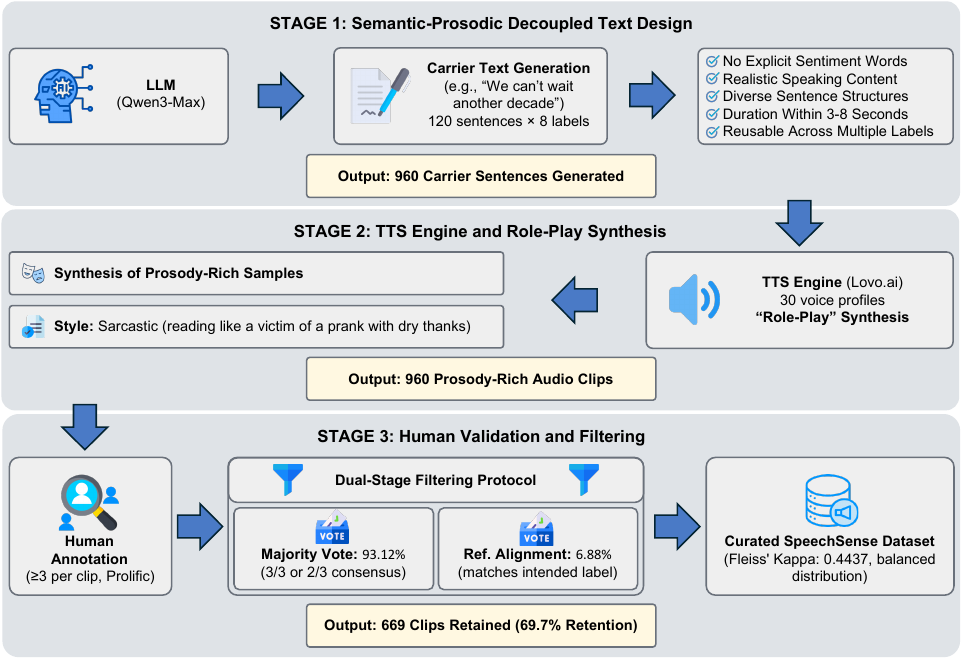}
\caption{The SpeechSense dataset construction pipeline. The framework consists of (1) semantic-prosodic decoupled text design, (2) role-play synthesis using Lovo.ai, and (3) dual-stage human validation and filtering.}
\label{fig:pipeline}
\end{figure}

We construct the dataset using a rigorous three-stage pipeline designed to minimize artifacts and maximize attitudinal distinctiveness (illustrated in Fig. \ref{fig:pipeline}).

\textit{1) Stage 1: Semantic-Prosodic Decoupled Text Design.} To ensure the model relies on acoustic cues rather than lexical priors, we generate \textit{carrier text}---semantically neutral sentences that are decoupled from the target labels. We utilize Qwen3-Max to generate 120 carrier sentences per label. The model plays a deliberately limited role here, producing only the carrier text fed to the TTS engine, while the attitude itself is realized at the synthesis stage. What matters for this role is therefore not reasoning ability but constraint adherence and lexical diversity, on which Qwen3-Max outperformed smaller-scale alternatives in our preliminary inspection.
We adhere to five strict criteria during generation: (1) exclusion of explicit emotional words; (2) realistic conversational content; (3) semantic neutrality to allow reuse across labels; (4) structural diversity (declarative, imperative, conditional) to prevent syntactic overfitting; and (5) moderate duration (3--8 seconds). This produces a corpus where the ``sentiment'' is conveyed purely by prosody.

\textit{2) Stage 2: TTS Engine Selection and Role-Play Synthesis.} 
Authentic fine-grained attitudinal speech is severely limited by data scarcity, privacy, and ethical constraints \cite{schuhmannEmoNetVoiceFineGrainedExpertVerified2025}, and recent benchmarks such as EmoNet-Voice \cite{schuhmannEmoNetVoiceFineGrainedExpertVerified2025} and VoxEval \cite{cui2025voxeval} likewise rely on commercial TTS engines. Synthesis further enables strict variable control: holding textual content constant across attitudes isolates prosody as the sole differentiating factor.
After evaluating six TTS systems including open-source (CosyVoice, IndexTTS2, Kokoro) and proprietary (GPT-4o, ElevenLabs) platforms, we select Lovo.ai\footnote{https://lovo.ai/} for its human-like prosody in rendering subtle non-basic sentiment cues, speaker diversity (30+ voice profiles), and scalable API infrastructure.

Crucially, to operationalize these capabilities, we implement a ``Role-Play'' synthesis methodology grounded in the Stanislavski method \cite{scherer2003vocal}. Rather than mechanically adjusting pitch parameters, we map our labels to specific situational acting directives supported by the engine (e.g., configuring the style for \textit{Sarcastic} as \textit{``reading like a victim of a prank with dry thanks''}). This approach leverages the engine's explicit style controls to yield organic, non-linear prosodic variations that mirror authentic human expression.


\textit{3) Stage 3: Human Validation and Filtering.} To rigorously filter out synthesis artifacts and expressive misalignments, we establish a robust human-in-the-loop validation pipeline. We deploy the evaluation task via Qualtrics\footnote{https://www.qualtrics.com/} and recruit qualified annotators through the Prolific\footnote{https://www.prolific.com/} platform, enforcing strict screening criteria on language proficiency, education, and historical approval rate (see Table~\ref{tab:annotation_stats} for full recruitment and filtering statistics).

During the annotation phase, each audio clip is evaluated by at least three independent annotators. We apply a dual-stage filtering protocol to curate the final dataset. First, under a Majority Vote criterion, clips achieving consensus (3/3) or majority agreement (2/3) are automatically retained. Second, we employ a Reference Alignment strategy for ambiguous cases where annotators lack consensus (i.e., no majority): clips are retained if at least one annotator matches the \textit{intended} label (the TTS generation target). This step preserves subtle but accurate expressions that may be perceived as ambiguous by some lay annotators, provided they align with the ground truth generation intent. This filtering strategy has been validated in previous crowdsourcing studies to maintain ground-truth quality \cite{smith2018crowdsourcing}. Crucially, a substantial 93.12\% of the final test samples are retained strictly through majority vote, indicating that the ground truth is predominantly driven by spontaneous human consensus rather than generation bias.

The final curated test dataset consists of 669 high-quality clips, achieving an overall Fleiss' Kappa of 0.4437. Based on the widely accepted interpretation guidelines established by Landis and Koch \cite{landisMeasurementObserverAgreement1977}, this score signifies moderate agreement (defined as the $0.41$--$0.60$ interval). 
This score is consistent with established benchmarks (CREMA-D: $\kappa=0.42$ \cite{caoCREMADCrowdSourcedEmotional2014}, IEMOCAP: $\kappa=0.40$ \cite{bussoIEMOCAPInteractiveEmotional2008}) and substantially exceeds recent fine-grained synthetic benchmarks (EmoNet-Voice: Krippendorff's $\alpha=0.14$ \cite{krippendorff2011computing, schuhmannEmoNetVoiceFineGrainedExpertVerified2025}).

\begin{table}[t]
\caption{Human Annotation and Filtering Statistics.}
\label{tab:annotation_stats}
\begin{tabular}{ll}
\toprule
\textbf{Metric} & \textbf{Value} \\
\midrule
\multicolumn{2}{l}{\textit{Annotator Recruitment}} \\
Platform & Prolific \\
Eligible candidate pool & 23,006 (from 150,000+) \\
Screening: Language & Native English \\
Screening: Education & Bachelor's+ \\
Screening: Approval rate & $\geq$99\% \\
Annotators per clip & $\geq$ 3 \\
\midrule
\multicolumn{2}{l}{\textit{Filtering Results}} \\
Pre-filtering clips & 960 \\
Stage 1: Majority Vote retained & 623 (93.12\% of final set) \\
Stage 2: Ref. Alignment retained & 46 (6.88\% of final set) \\
Final curated clips & 669 (69.7\% retention) \\
\midrule
\multicolumn{2}{l}{\textit{Inter-Annotator Agreement}} \\
Fleiss' Kappa (overall) & 0.4437 (Moderate) \\
\midrule
\multicolumn{2}{l}{\textit{Speaker-Label Balance}} \\
Speakers covering all 8 labels & 26 / 30 \\
Speakers covering 7 labels & 4 / 30 \\
\bottomrule
\end{tabular}
\end{table}

\subsection{Final Dataset Statistics}
\label{sec:dataset_statistics}

\begin{figure}[tb]
\centering
\includegraphics[width=0.95\columnwidth]{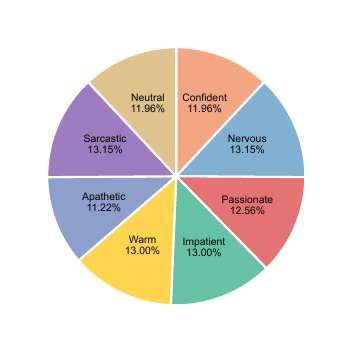}
\caption{SpeechSense Dataset Label Distribution.}
\label{fig:pie_chart}
\end{figure}

The pipeline yields two distinct data subsets. To ensure robust generalization and prevent downstream models from overfitting to specific lexical patterns, we employ distinct text generation sources for the training and evaluation partitions. The SpeechSense Test Set serves as the evaluation ``Gold Standard'', comprising 669 high-quality audio clips derived from Qwen3-Max carrier text. Following the rigorous human validation described in Section~\ref{sec:pipeline}, this subset maintains a meticulously balanced distribution across the eight classes, as visualized in Fig. \ref{fig:pie_chart}, with prevalence ranging from 11.2\% (\textit{Apathetic}) to 13.1\% (\textit{Nervous} and \textit{Sarcastic}). This balance minimizes class bias, ensuring that evaluation metrics reflect genuine prosodic understanding rather than prior probability shifts.
Complementing the evaluation set, the SpeechSense Training Set constitutes a larger corpus of synthetic data generated using a different LLM family (Gemini 3 Pro). By decoupling the text generation source between training and testing, we ensure that the dataset evaluates the model's ability to capture acoustic representations rather than specific writing styles or artifacts unique to a single LLM. Although both subsets share the same 30 voice profiles, this overlap does not introduce speaker identity leakage. We ensure a balanced speaker-label matrix: 26 of 30 speakers cover all 8 attitudes, and the remaining 4 cover 7 (see Table~\ref{tab:annotation_stats}). Consequently, speaker identity provides no predictive shortcut for the target label, and the model must genuinely disentangle fine-grained prosodic variations rather than memorize speaker timbre. Beyond label coverage, the 30 profiles also exhibit substantial acoustic diversity. Table~\ref{tab:dataset_speaker_stats} summarizes their variation measured on a shared control utterance, spanning a broad pitch range and distinct vocal tract characteristics. While this training set uses the same Role-Play TTS methodology to ensure prosodic consistency, it relies on the model's inherent stability for large-scale weakly-supervised learning. To ensure reproducibility, we publicly release the full dataset (audio, labels, and synthesis directives) under a CC BY 4.0 license, along with supplementary analyses, at \url{https://github.com/Sher13cked/SpeechSense}.



\begin{table}[tb]
\centering
\caption{SpeechSense Dataset and Speaker Statistics.}
\label{tab:dataset_speaker_stats}
\footnotesize
\setlength{\tabcolsep}{5pt}
\begin{tabular}{lcccc}
\toprule
\textbf{Dataset} & \textbf{Source} & \textbf{Val. Method} & \textbf{Clips} & \textbf{Spkrs} \\
\midrule
Training Set & Gemini 3 Pro & Weakly-Sup. & 1,522 & 30 \\
Test Set & Qwen3-Max & Human-Val. & 669 & 30 \\
\midrule
\multicolumn{5}{l}{\textit{Speaker Diversity} (shared control utterance, 30 profiles)} \\
\multicolumn{2}{l}{Gender balance} & \multicolumn{3}{l}{16 M / 14 F} \\
\multicolumn{2}{l}{Mean F0 range} & \multicolumn{3}{l}{190.6--315.8 Hz (std = 41.7)} \\
\multicolumn{2}{l}{Formant F1 / F2 (std)} & \multicolumn{3}{l}{58.3 / 94.2 Hz} \\
\bottomrule
\end{tabular}
\end{table}

\section{Experiments and Results}
\label{sec:experiments}

\subsection{Speech Synthesis Quality Validation}
\label{sec:quality_validation}

Given that prosodic variations (e.g., mumbling in \textit{Apathetic} or rapid speech in \textit{Impatient}) can inadvertently degrade speech clarity, it is crucial to ensure that the synthesized audio retains high linguistic content preservation.

We employ Whisper-large-v3 \cite{radford2023robust}, an advanced ASR model, to transcribe the entire SpeechSense benchmark. We compute the WER by comparing the ASR transcriptions against the ground-truth carrier text used for generation. 

\begin{table}[tb]
\centering
\caption{Word Error Rate Analysis.}
\label{tab:wer_analysis}
\resizebox{\columnwidth}{!}{%
\begin{tabular}{lccccccccc}
\toprule
\textbf{Dataset} & \textbf{Conf.} & \textbf{Nerv.} & \textbf{Warm} & \textbf{Apat.} & \textbf{Pass.} & \textbf{Impat.} & \textbf{Sarc.} & \textbf{Neut.} & \textbf{Avg.} \\
\midrule
\textbf{Training Set ($\downarrow$)} & 5.00\% & 2.32\% & 0.18\% & 9.31\% & 1.15\% & 6.18\% & 8.57\% & 9.98\% & \textbf{5.42\%} \\
\textbf{Test Set ($\downarrow$)}     & 2.90\% & 1.43\% & 3.88\% & 1.38\% & 7.20\% & 1.27\% & 7.98\% & 3.26\% & \textbf{3.70\%} \\
\bottomrule
\end{tabular}%
}
\end{table}

As presented in Table~\ref{tab:wer_analysis}, the benchmark achieves an overall WER of 3.70\% on the curated Test Set and 5.42\% on the weakly-supervised Training Set. This verifies that our TTS pipeline yields clear, artifact-free audio suitable for downstream representation learning.

A closer examination reveals performance nuances that align with our prosodic dimensions. The \textit{Sarcastic} class consistently exhibits higher WER (Test: 7.98\%, Train: 8.57\%), a pattern that is both expected and desirable, reflecting the acoustic complexity of sarcasm, which naturally challenges ASR models optimized for neutral speech. These results confirm that SpeechSense audio is of sufficient quality for the subsequent classification experiments.

\subsection{Experimental Setup}

We design our experimental protocol to answer two core questions: (1) whether acoustic cues are the dominant signal for fine-grained sentiment detection, and (2) whether the learned representations generalize across model architectures. To this end, we benchmark three model families under a unified evaluation framework.

\textit{1) Model Architecture.} As multi-modal LLMs, we employ Qwen2.5-Omni (3B and 7B), freezing the backbone and appending a linear classification head (Dropout $p=0.1$) to map final hidden states to 8 classes. As text-only baselines, we include Qwen2.5-Instruct (3B and 7B), which share 
the same language backbone but lack an audio encoder. As speech 
encoders, we evaluate Whisper-large-v3 \cite{radford2023robust}, 
HuBERT-large \cite{hsu2021hubert}, and Wav2Vec2-large 
\cite{baevski2020wav2vec}, each paired with attention pooling and a linear classification head. This design spans three paradigms---multi-modal understanding, text-only reasoning, and pure acoustic representation. All models share a unified evaluation protocol (frozen backbone + linear head + cross-entropy loss), ensuring that performance differences reflect representational capacity rather than training methodology.

\textit{2) Training Protocol.} All models follow the cross-source evaluation protocol defined in Section~\ref{sec:dataset_statistics}. For multi-modal LLMs and text-only LLMs, we apply Low-Rank Adaptation (LoRA) on the attention modules with the backbone frozen. For speech encoders, we adopt a two-stage strategy: first training only the classification head, then unfreezing the entire network for joint fine-tuning. Hyperparameters are listed in our supplementary repository.

\subsection{Main Results: The Primacy of Prosody}

Table~\ref{tab:main_results} summarizes performance across modalities, model families, and scales, comparing zero-shot and supervised settings. 

\begin{table}[tb]
    \centering
    \caption{Main Classification Results.}
    \label{tab:main_results}
    \resizebox{\columnwidth}{!}{%
    \begin{tabular}{llcccc}
        \toprule
        \multirow{2}{*}{\textbf{Model}} & \multirow{2}{*}{\textbf{Modality}} & \multicolumn{2}{c}{\textbf{Zero-shot}} & \multicolumn{2}{c}{\textbf{Supervised (Ours)}} \\
        \cmidrule(lr){3-4} \cmidrule(lr){5-6}
         &  & \textbf{Acc} & \textbf{Macro F1} & \textbf{Acc} & \textbf{Macro F1} \\
        \midrule
        \multicolumn{6}{l}{\textit{Multi-modal LLMs}} \\
        Qwen2.5-Omni-3B & Text & 8.22\% & 3.79\% & 25.26\% & 19.71\% \\
        Qwen2.5-Omni-3B & Audio & 3.74\% & 1.31\% & 54.86\% & 53.38\% \\
        Qwen2.5-Omni-7B & Text & 11.36\% & 6.20\% & 26.76\% & 22.27\% \\
        Qwen2.5-Omni-7B & Audio & 11.96\% & 2.96\% & \textbf{56.95}\% & \textbf{56.76\%} \\
        \midrule
        \multicolumn{6}{l}{\textit{Text-only LLMs}} \\
        Qwen2.5-Instruct-3B & Text & 15.84\% & 6.63\% & 14.20\% & 4.60\% \\
        Qwen2.5-Instruct-7B & Text & 11.36\% & 4.33\% & 25.26\% & 15.97\% \\
        \midrule
        \multicolumn{6}{l}{\textit{Speech Encoders}} \\
        Whisper-large-v3 & Audio & 13.30\% & 5.03\% & 45.44\% & 45.06\% \\
        HuBERT-large & Audio & 10.16\% & 3.87\% & 44.39\% & 43.79\% \\
        Wav2Vec2-large & Audio & 12.56\% & 3.17\% & 44.54\% & 42.45\% \\
        \bottomrule
    \end{tabular}%
    }
\end{table}

First, the results underscore the critical necessity of specific prosodic training. Across all model families, zero-shot performance remains near-random, with Macro-F1 scores ranging from 1.31\% to 6.63\%. This confirms that neither pre-trained multi-modal LLMs nor speech encoders inherently possess fine-grained attitudinal representations. However, fine-tuning on SpeechSense yields substantial performance gains across all architectures: multi-modal LLM audio models improve by up to 50 percentage points in accuracy, and speech encoders achieve 42--45\% Macro-F1 from near-random baselines. These improvements demonstrate that the dataset effectively teaches diverse architectures to decode complex prosodic cues that were previously inaccessible to their pre-trained backbones.

Second, the experiments empirically validate the semantic neutrality of our dataset design through converging evidence from two distinct text-only model families. After supervised training, Qwen2.5-Omni text models saturate at 20--22\% F1, and Qwen2.5-Instruct text models perform even worse---as low as 4.60\% F1 for the 3B variant, which is lower than its own zero-shot baseline (6.63\%), indicating that training induces mode collapse in the absence of acoustic signal. This consistent failure across architectures confirms that the textual content is indeed semantically neutral and insufficient for distinguishing attitudes. The significant performance gap between the best audio model (56.95\% Acc) and the best text model (26.76\% Acc) confirms that the task is solvable almost exclusively through the \textit{``how''} (paralinguistic information) rather than the \textit{``what''} (linguistic information).

Third, speech encoders validate that acoustic features alone carry substantial attitudinal signal even without language modeling capacity. Whisper, HuBERT, and Wav2Vec2 achieve 42--45\% Macro-F1, demonstrating that fine-grained prosodic cues can be captured by dedicated speech representations. The 10--14 percentage point gap between multi-modal LLM audio models (53--57\% F1) and speech encoders (42--45\% F1) suggests that the language understanding component of multi-modal LLMs provides additional reasoning capacity on top of prosodic features---likely by integrating acoustic evidence with contextual semantic inference.

\subsection{Fine-Grained Analysis and Failure Modes}

\textit{1) Models with Acoustic Access.} We analyze the class-wise performance of multi-modal LLMs (audio mode) and speech encoders---the two model families that receive acoustic input. Full per-class F1 scores are provided in our supplementary repository.

Across all audio models, \textit{Nervous} emerges as the most reliably detected attitude (68--80\% F1), suggesting that the acoustic markers of anxiety---pitch jitter, elevated mean pitch, and irregular rhythm---are robustly captured by speech representations alone, with speech encoders matching or surpassing multi-modal LLMs. In contrast, \textit{Neutral} and \textit{Confident} are consistently the hardest classes (below 45\% F1 for most models), reflecting the difficulty of detecting attitudes defined by the absence of distinctive prosodic features. The two model families also diverge systematically: for high-energy attitudes requiring valence disambiguation, such as \textit{Impatient} and \textit{Passionate}, multi-modal LLMs substantially outperform speech encoders, as these attitudes share similar energy profiles but differ in valence---a distinction that benefits from language understanding capacity.
Finally, the confusion structure indicates that residual errors are systematic rather than random. Across all audio models, misclassifications concentrate on a small number of prosodically adjacent pairs: \textit{Confident} and \textit{Neutral} are mutually confused in both directions, consistent with their shared lack of distinctive prosodic excursions; \textit{Confident}--\textit{Passionate} overlap in multi-modal LLMs through their common assertive intensity; and \textit{Warm}--\textit{Sarcastic} overlap in speech encoders, reflecting the exaggerated pleasantness characteristic of sarcastic delivery. That errors fall on acoustically interpretable neighbors rather than dispersing arbitrarily indicates genuine prosodic proximity and the difficulty of fine-grained stance discrimination, not ill-defined category boundaries. Full confusion matrices are provided in our supplementary repository.

\textit{2) Models without Acoustic Access.} We examine the text-only models---multi-modal LLMs (text mode) and text-only LLMs---which receive only transcripts. 
All four text-only configurations suffer from severe mode collapse, but critically, each collapses toward a \textit{different} dominant class: Omni-3B concentrates nearly 70\% of predictions on \textit{Impatient}, Omni-7B assigns the majority to \textit{Sarcastic}, Instruct-3B directs 87\% to \textit{Warm}, and Instruct-7B spreads across \textit{Confident}, \textit{Sarcastic}, and \textit{Impatient}. Meanwhile, several classes receive \textit{zero} predictions across multiple models---for instance, \textit{Neutral} is never predicted by any of the four text-only configurations. This inconsistency in collapse targets is itself revealing: if the textual content carried systematic attitudinal signal, different models should converge toward similar predictions. Instead, each model resorts to arbitrary priors that vary with architecture and scale, providing strong evidence that the text carries no learnable sentiment information. Detailed per-model prediction distributions are provided in our supplementary repository.

\subsection{Scaling and Architecture Effects}


For multi-modal LLM audio models, scaling from 3B to 7B yields a moderate gain (Macro-F1: 53.38\% $\to$ 56.76\%). This improvement is concentrated on cognitively complex classes such as \textit{Sarcastic} and \textit{Confident}, which require disentangling subtle or contradictory prosodic signals, while several other classes show no gain or slight regression. This suggests that the additional reasoning capacity of larger models primarily benefits specific classes rather than uniformly improving detection.


For text-only models, even the 7B variant reaches only 15.97\% F1, far below every audio configuration, confirming that scaling cannot compensate for the fundamental absence of acoustic signal.

Among speech encoders, Whisper (45.06\%), HuBERT (43.79\%), and Wav2Vec2 (42.45\%) converge within a narrow 2.6-point band despite distinct pre-training paradigms. This convergence suggests that the attitudinal signal in SpeechSense is robustly accessible to diverse acoustic representations and does not depend on any particular pre-training strategy.

\section{Limitations}
\label{sec:limitations}

We acknowledge several limitations. First, SpeechSense is built entirely from synthesized speech. While we validate prosodic fidelity through human annotation and WER analysis, a domain gap may exist between synthetic and real speech. We position SpeechSense as a foundational cold-start resource and encourage future work on real-world adaptation. Second, the current taxonomy covers eight interpersonal stances in English only. Extending to additional attitudes and languages would broaden applicability. Third, speaker diversity is limited to 30 voice profiles from a single TTS engine. Although our speaker-label balance analysis confirms no identity leakage, future iterations could incorporate multiple engines and a larger speaker pool to further improve generalization.

\section{Conclusion}
\label{sec:conclusion}
We introduce SpeechSense, a robust dataset that addresses the limitations of existing emotion recognition benchmarks by shifting the focus from basic emotions to fine-grained interpersonal stances, while rigorously isolating paralinguistic attitude from semantic content. We define a taxonomy of eight specific sentiment states, and employ high-fidelity synthesis to build a resource where acoustic generalization is critical. Our experiments across multi-modal LLMs, text-only LLMs, and speech encoders consistently demonstrate that acoustic access is necessary for detecting these nuanced distinctions, as semantic cues alone prove insufficient regardless of model scale. SpeechSense serves as a pivotal step toward more naturalistic human–machine interaction, enabling future models to understand not just the message, but also the social dynamics and nuanced delivery that shape the spoken word. 

\begin{acks}
The research presented in this paper was partially supported by the Research Grants Council of the Hong Kong Special Administrative Region, China (CUHK 2300246, RGC C1043-24G) and (CUHK 14203425, RGC GRF 2151317).
\end{acks}



\end{document}